\documentclass{article}

\usepackage{arxiv}

\usepackage[utf8]{inputenc} 
\usepackage[T1]{fontenc}    
\usepackage{hyperref}       
\usepackage{url}            
\usepackage{booktabs}       
\usepackage{amsfonts}       
\usepackage{nicefrac}       
\usepackage{microtype}      
\usepackage{lipsum}		
\usepackage{graphicx}
\usepackage{natbib}
\usepackage{doi}

\usepackage{amsmath,amssymb,mathtools}
\usepackage{algorithmic,algorithm}
\usepackage{amsthm}

\theoremstyle{definition}
\newtheorem{definition}{Definition}

\theoremstyle{plain}
\newtheorem{proposition}{Proposition}

\title{Enterprise Representation Simplification (ERS): Reducing Representational Complexity for Enterprise AI}

\author{ \href{https://orcid.org/0009-0000-9405-4795}{\includegraphics[scale=0.06]{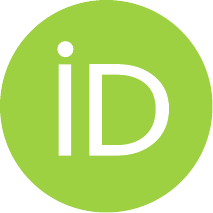}\hspace{1mm}Terry Dorsey}\\
	Denodo Corporation\\
	Palo Alto, California 94301 \\
	\texttt{tdorsey@denodo.com} \\
	\And
	\href{https://orcid.org/0009-0008-3989-4677}{\includegraphics[scale=0.06]{orcid.pdf}\hspace{1mm}Kevin Huggins} \\
	Harrisburg University\\
	Harrisburg, Pennsylvania 17101 \\
	\texttt{khuggins@harrisburgu.edu} \\
}

\renewcommand{\shorttitle}{Enterprise Representation Simplification (ERS)}

\hypersetup{
pdftitle={Enterprise_Representation_Simplification},
pdfsubject={},
pdfauthor={Terry Dorsey, Kevin Huggins},
pdfkeywords={Generative AI, Context Management,Data Architecture}
}

\begin{document}

\maketitle

\begin{abstract}
  Enterprise information is commonly represented through artifacts informed by applications, projects, technologies, organizational boundaries, and local implementation requirements. These structures accumulate over time, producing representational complexity that must be maintained by the enterprise and interpreted by information consumers, including artificial intelligence systems. This paper introduces Enterprise Representation Simplification (ERS) as the reduction of unnecessary representational complexity while preserving the information required within a defined scope. It then introduces Enterprise Representation Complexity (ERC) as a representation-neutral model for measuring and comparing such complexity across heterogeneous representation states.
  
  ERC characterizes representational extent through four dimensions: Representation Objects, Interactions, Behaviors, and Supporting Sources. Objects, Interactions, and Behaviors form dependent representation categories, while Supporting Sources characterize the extent through which the representation is exposed. ERC is defined at both representation and task levels, enabling comparison of enterprise representations independently of implementation technology and distinguishing architectural simplification from retrieval optimization.
  
  The paper further develops two consequences of Enterprise Representation Simplification. First, representational structures create lifecycle obligations associated with maintenance, governance, dependency management, change, enhancement, and operation. The paper derives an economic model that distinguishes recurring global representation cost, recurring task-level cost, and the one-time cost of transforming between representation states, enabling simplification to be evaluated over a defined time horizon. Second, reductions in task-level ERC reduce the representational extent that an AI system must identify, relate, and interpret. Prior Text-to-SQL research provides empirical support for the proposition that reduced schema and reasoning complexity can improve reasoning accuracy.
  
  ERC is not proposed as a universal complexity, performance, or cost metric. It provides measurable architectural variables through which representational alternatives, transformation effects, economic outcomes, and AI reasoning performance can be systematically compared and empirically evaluated.
\end{abstract}


\section[introduction]{Introduction}
Enterprise information environments rarely emerge from a single architectural design. They accumulate through applications, projects, acquisitions, technologies, organizational boundaries, and local requirements. Information concerning the same enterprise concepts may consequently be fragmented across tables, views, APIs, documents, events, reports, and application structures. Relationships may be represented through joins, mappings, interfaces, bridge structures, and other intermediary mechanisms, while business behavior may be distributed across transformations, queries, code, policies, calculations, and documentation.

The resulting complexity is not limited to physical implementation. It exists in the representation through which enterprise information is made available to consumers. Two representations may satisfy equivalent enterprise information requirements while differing substantially in the number of Objects, Interactions, Behaviors, and Sources that must be understood, governed, maintained, changed, and consumed.

The growing use of AI across the enterprise makes this more important. Retrieval-Augmented Generation (RAG), GraphRAG, semantic retrieval, context engineering, tool use, and orchestration seek to identify and assemble information relevant to a reasoning task. These mechanisms can improve the selection of information from a complex representation, but they do not necessarily reduce the complexity of the representation from which that information must be selected. A sophisticated retrieval mechanism may therefore compensate for representational complexity without addressing whether that complexity is necessary.

This paper addresses an upstream architectural question: Can enterprise information be represented more simply while preserving the information required by the enterprise?

ERS is introduced as an information-preserving architectural transformation intended to reduce unnecessary representational complexity within a defined enterprise information scope. Enterprise Representation Complexity (ERC) is introduced to make that transformation measurable.

ERC operationalizes representational complexity as representational extent and not every possible form of structural, semantic, computational, or organizational complexity. It measures the number of distinguishable Objects, Interactions, Behaviors, and Supporting Sources through which information within a defined scope is represented. This permits heterogeneous representations to be compared without assuming that they use the same technology or modeling form.

The paper makes six contributions, some introducing new constructs and others extending or synthesizing established architectural principles within a formal theory of Enterprise Representation Simplification. First, it defines ERS independently of implementation technology and establishes information sufficiency as the preservation boundary for simplification. Second, it characterizes enterprise representation through Objects, Interactions, Behaviors, and Supporting Sources and establishes design principles for separating representational concerns, generalizing shared foundations while preserving necessary distinction at the appropriate scope, and identifying where simplification creates value. Third, it introduces ERC as a four-dimensional, representation-neutral measure of representational extent. Fourth, it distinguishes representation-level ERC from the minimal sufficient representational extent required for an individual task. Fifth, it formalizes representation-state change and distinguishes simplification from extension and other transformations. Sixth, it establishes a basis for relating changes in representational extent to enterprise obligations and for empirically evaluating their relationship to AI task accuracy, representation-dependent task cost, and the expected resource and economic cost of successful task outcomes.

These contributions do not depend on abstraction, generalization, separation of concerns, modularity, or other established architectural principles being individually novel. ERS applies and extends these principles at the level of enterprise representation, with information sufficiency defining the boundary within which simplification occurs. Its design principles address not only reduction in representational extent, but the placement of shared meaning and necessary distinction within that representation. ERC provides the corresponding measurement model. ERS does not assume that semantic enrichment or structural reduction is independently optimal; it treats unnecessary representational complexity as the condition to be reduced while preserving the information required for the defined scope.

Beyond evaluating individual representation transformations, ERC provides a basis for cumulative organizational learning. By repeatedly observing how changes in individual ERC dimensions correspond to changes in AI accuracy, representation-dependent task cost, lifecycle obligations, and enhancement effort, an enterprise can begin to identify which forms of representational change produce the greatest value in its own environment. These relationships may then inform resource allocation, architectural priorities, and future representation design. ERS can therefore be treated not only as the goal of an individual transformation, but as a continuing architectural goal supported by measurement, simplification, empirical evaluation, and improvement over time.

\section[relatedWork]{Related Work}
ERS intersects several bodies of research that address enterprise representation, architectural complexity, semantic interpretation, and AI consumption. These areas have generally developed independently: enterprise and semantic modeling focus on how information and meaning are represented; software and systems architecture address how complexity, dependency, and change are managed; and recent AI research examines how representation and context affect reasoning performance. ERS draws these bodies of research together by treating enterprise representation itself as an architectural object that can be simplified, measured, and evaluated relative to a defined information scope.
\subsection{Enterprise and Semantic Representation}
Enterprise architecture and conceptual modeling have long sought to provide abstractions through which complex organizational structures, information, processes, and relationships can be understood independently of their physical implementation. Conceptual modeling, in particular, captures and represents relevant aspects of a domain at a level of abstraction that supports understanding, communication, and design (\cite{anaby-tavor_insights_2010, rajabi_survey_2026, storey_conceptual_2023}). Established conceptual-modeling principles include classification, generalization and specialization, and aggregation and decomposition (\cite{atkinson_model-driven_2003}).

Semantic approaches extend this objective by describing enterprise information in terms of its business meaning. Ontologies formalize concepts, relationships, and constraints to support shared interpretation and interoperability (\cite{noy_ontology_2001, opdahl_ontology_2012, rajabi_survey_2026,song_ontology-driven_2013}). Knowledge graphs similarly organize information around entities and their relationships, providing a conceptual representation that connects information across enterprise sources and supports machine reasoning (\cite{hogan_knowledge_2022,ji_survey_2022, sellami_knowledge_2025}).

Semantic layers serve a related purpose by separating business-facing concepts, measures, and rules from underlying physical data structures. Recent AI research provides evidence that this explicit business context can materially affect reasoning performance (\cite{kim_semantic-layer-mediated_2026,rumiantsau_semantic_2026}). \cite{rumiantsau_semantic_2026} compare schema-only prompting with the same warehouse schema augmented by a compact semantic document describing measures, conventions, and disambiguation rules. Across three frontier models, the semantic condition improves accuracy by approximately 17–23 percentage points (\cite{rumiantsau_semantic_2026}). 

These approaches establish the importance of explicit enterprise meaning, but they address a different primary objective from ERS. Ontologies, knowledge graphs, conceptual models, and semantic layers provide mechanisms for representing, organizing, or enriching meaning; they do not inherently determine the minimum representation necessary for a defined enterprise information scope. Such representations may themselves be modular, extensible, or simplified. ERS therefore complements them by asking whether representation can be consolidated, generalized, abstracted, separated, or eliminated while preserving required information.

\subsection{Architectural Management of Complexity and Change}
A second body of research addresses complexity through architectural organization. Principles including separation of concerns, abstraction, generalization, modularity, information hiding, and loose coupling seek to isolate responsibilities and reduce unnecessary dependencies so that portions of a system can be understood and changed without requiring corresponding change throughout the system.

\cite{parnas_criteria_1972} provides a foundational argument for decomposing systems according to design decisions likely to change, extending decomposition beyond the sequence of processing steps. Information hiding allows such decisions to remain localized behind comparatively stable interfaces (\cite{abgrall_schema_2024,parnas_criteria_1972,singh_transformation_2015}). These findings are relevant to ERS because they demonstrate how flexibility can be localized to limit the propagation of variation through dependent structures.

Separation of concerns provides a related basis for distinguishing different forms of representational responsibility (\cite{aversano_review_2015,daga_separation_2006,parnas_criteria_1972,noauthor_pdf_1985}). Generalization and abstraction similarly allow common characteristics to be represented at a reusable level while unnecessary implementation detail is hidden from representations that do not require it (Cuesta et al., 2006). Loose coupling extends this principle to dependencies, seeking to minimize the extent to which changes in one component require corresponding changes elsewhere (Cuesta et al., 2006).

Modularity research further connects separation and information hiding to flexibility and economic value. \cite{baldwin_design_2000} distinguish visible architectural design rules from design decisions that can remain hidden within modules and evolve with greater independence. Their economic analysis treats this flexibility as a form of option value: modular architectures create opportunities for substitution and experimentation that more tightly interdependent architectures do not provide (\cite{baldwin_design_2000,noauthor_pdf_1985,clark_option_2002}).

Collectively, this literature establishes an architectural basis for separating concerns, localizing variability, reducing unnecessary dependencies, and preserving stable boundaries around structures that should not change together. These principles are relevant to enterprise representation because different aspects of representation may vary independently and may require different approaches to definition, governance, maintenance, and evolution. They also suggest that flexibility can be located where variation can be accommodated without unnecessarily propagating through dependent structures. Where variability can be coherently represented at a shared level, more foundational representation and dependent consumers may remain comparatively stable. These principles provide part of the architectural foundation for ERS.

\subsection{Representation Complexity and AI Reasoning}
Recent AI research provides a third foundation by demonstrating that reasoning performance depends not only on whether required information is available but also on how much information must be considered and how that information is represented.

Long-context research demonstrates that simply providing more information does not ensure that a model will use it effectively. \cite{liu_lost_2024} find substantial performance differences depending on the position of relevant information within long contexts, demonstrating limitations in models' ability to reliably exploit increasing amounts of available context (\cite{liu_lost_2024}). 

Text-to-SQL provides particularly relevant evidence because enterprise information constitutes an explicit representation that AI systems must interpret. The BEAVER enterprise benchmark demonstrates that enterprise Text-to-SQL requires models to resolve multiple representational challenges, including identifying relevant information, determining relationships across information structures, mapping business concepts to represented attributes and values, incorporating domain knowledge, and resolving complex analytical requirements (\cite{chen_beaver_2026}). The substantial performance improvement observed when models are provided with explicit information about relevant information structures, relationships, concept and value mappings, and domain knowledge further suggests that interpreting enterprise representation constitutes a significant component of the reasoning problem  (\cite{chen_beaver_2026}).

Schema-linking research addresses this problem by reducing the representation presented to the model for an individual task. AutoLink, for example, identifies full-schema exposure as problematic because of context-window limitations and noise from irrelevant information and dynamically identifies a relevant subset of the existing schema for query generation (\cite{wang_autolink_2026}). Such approaches demonstrate the value of reducing task-facing complexity, but achieve this primarily by selecting relevant information from an existing representation rather than simplifying the persistent representation from which task-specific information is selected.

More direct evidence comes from research examining how alternative representations of database information affect Text-to-SQL performance. \cite{zhang_case_2026} evaluate semantics-preserving transformations involving schema abstraction, workload-aware partitioning, and descriptive renaming. Across multiple Text-to-SQL pipelines and language models, these transformations improve execution accuracy by up to 4.2\% (\cite{zhang_case_2026}). Because the transformations preserve the underlying information semantics while changing the representation presented to the model, the results provide direct evidence that representational organization can affect an AI system's ability to reason over otherwise equivalent information.

This result is particularly relevant to the present work. \cite{zhang_case_2026} establish within logical database design that information-equivalent representations need not impose equivalent reasoning demands on an AI system. Their findings provide empirical support for examining representation itself as a factor in AI reasoning and motivate extending this question beyond relational schema design to broader forms of enterprise representation.

\subsection{Representation Complexity, Economic and Resource Consequences}
The architectural literature provides a basis for expecting representational organization to have economic consequences. Modularity and information hiding reduce dependencies among components and can permit localized change, substitution, and independent evolution. \cite{baldwin_design_2000} further demonstrate that flexibility created by modular design can have economic value, while recognizing that establishing modular architecture itself imposes cost (\cite{baldwin_design_2000} ).

Technology and AI economics provide complementary perspectives. FinOps unit economics relates technology costs to measurable units of business activity or value (\cite{noauthor_capability_2026,keller_expanding_2026}). AI-agent research similarly demonstrates the importance of evaluating execution cost at the task level. \cite{kapoor_ai_2024} show that agent designs can differ substantially in cost for comparable levels of performance and argue for jointly evaluating and optimizing accuracy and cost. Subsequent work extends this perspective across agent models and agent scaffolds, which structure how models perform tasks (\cite{kapoor_holistic_2026}), while \cite{smekal_can_2026} demonstrates that differences in task specification can materially affect token consumption. Together, these studies establish that AI execution cost is not fixed by the task alone but can vary with the information provided to perform the task and the mechanisms through which it is executed.

Environmental resource consumption provides another potential consequence of representational structure. Recent research demonstrates that AI inference energy varies with workload characteristics, including reasoning and output requirements, while prompt and agent design can materially alter energy consumption for comparable tasks (\cite{han_power_2026,krupp_promoting_2025,oviedo_energy_2026,podder_green_2026}). These findings indicate that environmental cost is not exclusively a property of the underlying model or hardware; it may also depend on how information and computational work are organized for task execution.

This distinction is important for ERS. Representation is not assumed to be costly merely because it exists, nor is simplification assumed to produce economic benefit automatically. Representational structures may create development, maintenance, governance, security, integration, operational, computational, dependency-management, and change obligations whose magnitude depends on the enterprise environment.

Existing research therefore supports relationships among architectural structure, dependencies, task representation, and economic outcomes, but does not provide universal monetary weights for the representational dimensions defined by ERC. ERS consequently keeps measurement and economic consequence separate. ERC characterizes representational complexity, while observed costs can subsequently be associated with changes in individual ERC dimensions. Their marginal economic relationships can therefore be estimated empirically rather than assumed a priori.

\subsection{Synthesis and Research Gap}
Taken together, prior research establishes several important foundations. Enterprise modeling, ontologies, knowledge graphs, and semantic layers demonstrate the value of making enterprise meaning explicit. Architectural research establishes abstraction, generalization, separation of concerns, information hiding, modularity, and loose coupling as mechanisms for managing complexity, dependency, and change. AI research demonstrates that both semantic context and the complexity and organization of representation can materially affect reasoning performance. Economic research on modularity further demonstrates that architectural flexibility can itself have measurable value.

These bodies of work, however, leave a gap between representing enterprise meaning and measuring whether the representation itself contains unnecessary complexity. Semantic enrichment can add information needed for interpretation, while retrieval and schema-selection techniques can reduce the representation presented to an individual task. Neither necessarily reduces the underlying representation that the enterprise must maintain, govern, secure, evolve, and expose to consumers.

ERS addresses this gap by treating representational complexity as an architectural condition that can be deliberately transformed while preserving a defined information scope. It distinguishes Objects, Interactions, Behaviors, and Supporting Sources, applying established principles of abstraction, generalization, separation of concerns, loose coupling, and modularity to enterprise representation while retaining these dimensions separately for measurement. ERC provides a corresponding representation-neutral measure of representational extent.

ERC does not assume that all forms of representational complexity have equivalent consequences, but allows these dimensions to be observed independently and their effects examined empirically. These effects may include utilization, dependency, change propagation, lifecycle obligations, task economics, and AI reasoning performance. ERS therefore does not assume that simpler representation is universally preferable. It provides a basis for examining where representational complexity resides, what representation is necessary within a defined scope, where flexibility should be located, and whether simplification produces measurable effects.

This perspective also extends to task execution. Different representations of equivalent task-relevant information may impose different computational obligations, including source access, information retrieval, relationship resolution, context construction, inference, tool invocation, and validation. Representations that preserve equivalent information for a task may therefore differ in the computational resources required to perform that task.

\section[EnterpriseRepresentation]{Enterprise Representation}
\subsection{Representation Categories}
Enterprise information can be characterized through three dependent representation categories: Representation Objects, Representation Interactions, and Representation Behaviors.
A \textbf{Representation Object} is a distinguishable structure through which information about something is maintained or exposed within the representation being considered. Depending upon the representation, an Object may correspond to a table, view, API resource, document-defined structure, application object, logical object, semantic object, or higher-order enterprise abstraction.

A \textbf{Representation Interaction} is a distinguishable representation of a relationship or action among Representation Objects.
A \textbf{Representation Behavior} is a distinguishable representation of logic that governs, derives, constrains, interprets, or operates upon Objects or their Interactions. Behaviors may include calculations, rules, policies, constraints, metrics, classifications, or other operational semantics.

Together, the categories distinguish what is represented, how represented things relate or interact, and what logic governs or derives them. They describe representational roles and thus do not dictate specific implementation artifact types. Heterogeneous artifacts can therefore be classified according to the role they perform within a representation independent of the technology through which they are implemented.
The categories have a dependency ordering:

\begin{equation}
    O \to I \to B
    \label{eq:placeholder_label}
\end{equation}
Interactions depend upon participating Objects, while Behaviors may depend upon Objects, Interactions, or both. This ordering denotes representational dependency and does not imply relative importance, cost, or numerical weight.

Abstraction may occur independently within each category and at different levels. Implementation-specific Objects may be transformed into enterprise Objects; implementation relationships may become enterprise Interactions; and duplicated, distributed, or implicit logic may become explicit enterprise Behaviors. The resulting representation need not impose the same level of abstraction across all three categories.

\textbf{Artifact Boundaries}
A single implementation artifact may contain multiple representation elements. Classification is based on what the artifact represents, not its technical form. For example, a view may represent an Object while also containing calculations. Calculations that derive an attribute of the Object are treated as part of that Object, while calculations that express independently meaningful business rules, metrics, classifications, or policies are represented as Behaviors. Thus, ERC counts the enterprise representation expressed through an artifact rather than the artifact itself.

\subsection{Supporting Sources}
A \textbf{Supporting Source} is a distinguishable source through which some portion of an enterprise representation is exposed to its consumers.
Supporting Sources may include systems, databases, APIs, documents, event streams, reports, logical information services, or other exposed information resources. The classification is relative to the representation being considered. Underlying implementation artifacts that are encapsulated behind an exposed representation and are not themselves exposed to its consumers are not Supporting Sources of that representation.

Supporting Sources relate to Objects, Interactions, and Behaviors by characterizing the sources through which those representations are exposed to consumers. They do not constitute another level of the Object–Interaction–Behavior dependency hierarchy. Objects, Interactions, and Behaviors characterize the semantic extent of the representation, while Supporting Sources characterize its exposure extent.
\begin{equation}
  (O, I, B) \xrightarrow{\textit{exposed through}} S
  \label{eq:placeholder_label}
\end{equation}

This distinction allows representations with similar semantic extent but different exposure characteristics to be distinguished without incorporating implementation lineage or source-to-structure dependency topology into the representation model. Supporting Sources are therefore bounded by exposure: an underlying implementation artifact is included only when it is itself exposed as part of the representation available to consumers. This boundary preserves the distinction between representational extent and implementation complexity and allows ERC to remain independent of the particular implementation used to realize a representation.
Excluding underlying implementation artifacts from representational extent does not imply that implementation is independent of representation. 

Existing implementation may constrain or create dependencies for establishing or changing a representation, and a transformation of the representation may in turn require changes to its supporting implementation. The effort and cost associated with establishing or modifying that implementation are therefore relevant to the economic evaluation of a representation transformation, even though the underlying implementation artifacts are not themselves counted as Supporting Sources unless they are exposed within the representation.

\subsection{Theoretical Basis for Representational Separation}
The separation of Objects, Interactions, and Behaviors is motivated by recurring distinctions across conceptual modeling, enterprise architecture, knowledge representation, semantic modeling, and enterprise AI. Although these traditions use different terminology, they commonly distinguish among the things represented within a domain, the relationships among those things, and the rules, constraints, calculations, policies, or other semantics governing them. The Entity-Relationship model provides an early example through its separation of entities and relationships, while business-rule, ontology, semantic-modeling, and contemporary semantic-layer research further distinguish governing or derived semantics from the structures to which they apply (\cite{chen_entity-relationship_1976}).

These recurring distinctions support the Object–Interaction–Behavior separation adopted by ERS. Because these categories perform different representational roles, transformations may affect them differently. For example, consolidating distributed logic into an explicit Behavior may increase Behavior representation while reducing Objects or Interactions and localizing subsequent change. Maintaining these distinctions allows such changes to be characterized independently and subsequently examined in relation to outcomes such as reasoning accuracy, context-construction effort, validation, remediation, lifecycle obligations, and change effort.

Supporting Sources serve a different role. Objects, Interactions, and Behaviors characterize what is represented, while Supporting Sources characterize the sources through which that representation is made available. Separating these concerns allows changes in representation to be distinguished from changes in source exposure.

\section[ERC]{Enterprise Representation Complexity}
\subsection{Representation-Level ERC}
For representation R, let \(O(R)\), \(I(R)\), \(B(R)\), and \(S(R)\) denote the sets of distinguishable Representation Objects, Representation Interactions, Representation Behaviors, and Supporting Sources. ERC characterizes each category through a category-specific complexity function.
\begin{equation}
    \mathrm{ERC}_{L}(R) = \varphi_{L}\!\bigl(X_{L}(R)\bigr), \quad L \in \{O, I, B, S\}
    \label{eq:placeholder_label}
\end{equation}

where \(X_L (R)\) is the set associated with category \(L\) and \(\phi_L\) is the complexity function applied to that category. The category-specific formulation permits representational complexity to be operationalized differently where a justified measure is available, without assuming that complexity must be measured identically across categories.
\begin{definition}[Enterprise Representation Complexity]
\label{def:def_erc}
\begin{equation}
    \begin{split}
    \label{eq:placeholder_label}
    \operatorname{ERC}(R) = \bigl(&\phi_O\bigl(O(R)\bigr),\\ &\phi_I\bigl(I(R)\bigr),\\
    &\phi_B\bigl(B(R)\bigr),\\ 
    &\phi_S\bigl(S(R)\bigr)\bigr)        
    \end{split}
\end{equation}
\end{definition}
For comparative analysis, each ERC dimension must be measured consistently across the representation states being compared. For example, Objects must be measured using the same criteria in each representation, as must Interactions, Behaviors, and Supporting Sources. ERC remains multi-dimensional, and its components are not aggregated into a single score because the measure does not assume that units of Object, Interaction, Behavior, and Supporting Source complexity are equivalent.

\subsection{Cardinality as the Present Operationalization}
The ERC measurement does not require cardinality to be the only possible measure of representational complexity. For simplicity, reproducibility, and technology independence, however, the theoretical development and illustrative analysis in this paper operationalize each category-specific complexity function using cardinality:
\begin{equation}
    \phi_{L}(X) = \operatorname{card}(X)
    \label{eq:placeholder_label}
\end{equation}
Accordingly, for the present paper:

\begin{equation}
    \begin{split}
        \label{eq:placeholder_label}
        ERC(R) = \bigl(&\operatorname{card}(O(R)), \\
        &\operatorname{card}(I(R)), \\
        &\operatorname{card}(B(R)), \\
        &\operatorname{card}(S(R))\bigr)
        \end{split}
    \end{equation}        

Cardinality is directly observable and avoids introducing unsupported weighting assumptions. Under this operationalization, ERC characterizes representational size within each category rather than total semantic, cognitive, computational, economic, or organizational difficulty. An additional structure therefore contributes one unit to the applicable category count, but ERC does not assume that every structure imposes equal burden or produces equal consequences.

This distinction is especially useful when comparing information-equivalent representations. If \(R_a\)  and \(R_b\) preserve the information required by the same scope Q,
\begin{equation}
    R_a \equiv_{Q} R_b
    \label{eq:placeholder_label}
\end{equation} 
then differences in cardinality identify differences in the amount of distinguishable representation used to preserve that scope without assigning predetermined cost or value to each additional structure. Alternative complexity functions may later incorporate empirically justified structural or other characteristics without changing the underlying ERC metric.

\subsection{Task-Level ERC}
A task typically requires only a subset of the enterprise representation. For reasoning task q, let \(O(q,R)\), \(I(q,R)\), \(B(q,R)\), and \(S(q,R)\) denote the minimal sufficient sets of ERC dimensions required to satisfy \(q\).

\begin{definition}
\label{def:def_task_erc}
\textbf{(Task-Level Enterprise Representation Complexity)} \\
\begin{equation}
    \begin{split}
    ERC(q,R) = \bigl(&\phi_O\bigl(O(q,R)\bigr),\\ &\phi_I\bigl(I(q,R)\bigr),\\
    &\phi_B\bigl(B(q,R)\bigr),\\
    &\phi_S\bigl(S(q,R)\bigr)\bigr)
    \label{eq:placeholder_label}      
    \end{split}
\end{equation}
\end{definition}

Under the cardinality operationalization used in this paper:
\begin{equation}
    \begin{split}
        \label{eq:placeholder_label}
        ERC(R) = \bigl(&\operatorname{card}(O(q,R)), \\
        &\operatorname{card}(I(q,R)), \\
        &\operatorname{card}(B(q,R)), \\
        &\operatorname{card}(S(q,R))\bigr)
        \end{split}
    \end{equation}

The minimal-sufficiency condition distinguishes task-level ERC from retrieval behavior. A retrieval mechanism may select information beyond the minimum required for q; such over-selection characterizes retrieval performance rather than \(ERC(q,R)\). Under the cardinality operationalization, task-required sets are subsets of the available representation, giving:
\begin{equation}
\text{ERC}(q, R) \preccurlyeq \text{ERC}(R)
\end{equation}
where \(\preccurlyeq\) denotes component-wise inequality.

\subsection{Representation State Utilization and Coverage}
Enterprise Representation Complexity characterizes the extent of a representation independently of any particular workload. Workload observations provide a complementary view by characterizing how that representation is actually consumed. Three related measures are useful for this purpose: element utilization, category utilization, and category coverage. These describe the relationship between a representation state and an observed workload.

\subsubsection{Element Utilization}
Let \(W\) denote an observed workload and \(F_q\) the observed execution frequency of task \(q\). For representational element x, element utilization is:
\begin{equation}
U(x) = \frac{\sum_{q \in W} F_q \cdot \mathbf{1}[x \in R_q]}{\sum_{q \in W} F_q}
\end{equation}

where \( R_q\) is the minimal sufficient representation required for task \(q\), and \(\mathbf{1}[x \in R_q]\) equals one when x participates in \(R_q\) and zero otherwise.

Element utilization represents the proportion of observed task executions that require x. A value near one indicates that the element participates in most observed executions, whereas a value near zero indicates infrequent participation. Low observed utilization does not imply that an element is unnecessary; it may support infrequent, unobserved, future, regulatory, or otherwise required tasks.

\subsubsection{Category Utilization}
Element utilization can be extended to characterize how intensively an entire representation category is consumed. For category \(C\), let \(C\) denote the set of available elements in that category and \(C_q\) the subset required by task \(q\). Category utilization is:
\begin{equation}
U(C) = \frac{\sum_{q \in W} F_q |C_q|}{|C| \sum_{q \in W} F_q}
\end{equation}
Category utilization represents the average proportion of the available category required by observed task executions.

For the cardinality operationalization of ERC, \(C\) may represent Objects, Interactions, Behaviors, or Supporting Sources, and \(|C_q |\) and \(|C|\) correspond directly to the applicable task-level and representation-level ERC counts. Category utilization therefore connects the extent maintained in the representation with the extent typically required by the observed workload.

\subsubsection{Category Coverage}
Utilization captures frequency of participation, but it does not describe how broadly the available representation is employed. For category \(C\), category coverage is:
\begin{equation}
\text{Coverage}(C) = \frac{\left| \bigcup_{q \in W} C_q \right|}{|C|}
\end{equation}

Category coverage represents the proportion of the available category required by at least one task in the observed workload.
Unlike utilization, coverage is insensitive to execution frequency. An element used once and an element used repeatedly both contribute equally once they appear in the workload. Coverage therefore characterizes the breadth of demonstrated workload relevance, whereas utilization characterizes its intensity.

\subsubsection{Interpretation}
Together, the three measures provide complementary views of a representation state. Element utilization identifies where demand is concentrated; category utilization indicates how much of a category is typically required by an execution; and category coverage indicates how much of the category is exercised at all. Representations with identical ERC may therefore exhibit materially different workload-consumption profiles.

These measures are intentionally workload relative. They provide empirical observations that can support subsequent analysis of simplification opportunities, task-level representational demand, lifecycle obligations, and the relationship between representational extent and consumer outcomes.

\subsection{ERC Measurement}
ERC may be assessed manually, computationally, or through a combination of both. Measurement requires identifying representation elements within a defined scope and classifying them as Objects, Interactions, Behaviors, or Supporting Sources according to explicit classification criteria. ERC does not prescribe a particular analysis mechanism; implementations may use architectural analysis, metadata inspection, automated parsing, or AI-assisted classification. Reproducibility depends on making the representation boundary, classification rules, and treatment of ambiguous elements explicit.

\subsection{Utility of ERC}
ERC provides a common architectural measurement structure through which enterprise representations can be characterized, compared, and studied over time. At the representation level, it makes differences among architectural states explicit; at the task level, it characterizes the minimal representation required to satisfy individual information requirements; and across observed workloads, it supports analysis of representation utilization. Across repeated transformations and tasks, the separate ERC dimensions provide comparable architectural variables through which relationships with accuracy, task cost, validation, remediation, lifecycle obligations, and change effort can be empirically investigated.

ERC provides the measurement structure necessary to determine whether and how changes in enterprise representation are associated with them. The cardinality function used in this paper is therefore an initial operationalization and does not claim that representational complexity is intrinsically reducible to counts. Future empirical work may evaluate alternative category-specific complexity functions where they provide additional explanatory or decision-support value.

\section[EnterpriseRepresentationSimplification]{Enterprise Representation Simplification}

\subsection{Information Scope and Information Preservation}
Enterprise Representation Simplification is evaluated within a defined enterprise information scope Q. The scope establishes the enterprise information whose representation is being compared across representation states. Information preservation then requires that the information represented within that scope remain equivalent after simplification; preservation is not satisfied merely because a selected task or subset of requirements can still be answered.
Let \(\text{Info}_Q(R)\) denote the enterprise information represented by \(R\) within scope \(Q\).

\begin{definition}[(Information Preservation]
\label{def:def_informationpreservation}
A transformation \(T: R_0 \to R_1\) is information preserving with respect to \(Q \) when the enterprise information represented within \(Q\) is preserved across the transformation.
\begin{equation}
\text{Info}_Q(R_1) \equiv \text{Info}_Q(R_0)
\end{equation}
\end{definition}

Information preservation concerns equivalence of the scoped enterprise information rather than identity of representational structure.\( R_0\)  and \(R_1\) may differ substantially in the number, organization, or abstraction level of their representation artifacts while preserving the same enterprise information.

\subsection{Operational Evaluation of Information Preservation}
Let \(Q\) denote the set of information requirements within the defined scope:
\begin{equation}
Q = \{q_1, q_2, \dots, q_n\}
\end{equation}

For each requirement \(q \in Q\), a preservation criterion specifies what must remain available following transformation for that requirement to continue to be satisfied. Depending on the requirement, this may include information values, relationships, Behavior outcomes, historical distinctions, source provenance, or other necessary characteristics. Define \(\text{Preserve}(q, R_0, R_1)\) as an indicator of whether the applicable preservation criterion for \(q\) is satisfied:
\begin{equation}
\text{Preserve}(q, R_0, R_1) \in \{0, 1\}
\end{equation}
Representations \(R_0\) and \(R_1\) are operationally information-equivalent with respect to \(Q\) when every information requirement satisfies its applicable preservation criterion:
\begin{equation}
R_0 \equiv_Q R_1 \iff \forall q \in Q, \text{Preserve}(q, R_0, R_1) = 1
\end{equation}

The preservation test may therefore differ by requirement. It may involve equivalent query results, preservation of required attributes or relationships, equivalent Behavior outcomes, retention of required historical distinctions or source provenance, or documented domain validation. The measurement does not prescribe a universal equivalence test; the preservation criteria must be defined for the applicable requirements and applied consistently.

\subsection{Unnecessary Representational Complexity}
Unnecessary representational complexity is representational extent that can be eliminated without diminishing the enterprise information represented within \(Q\). It is therefore established through information-preserving reducibility and not through a subjective determination that an individual artifact is unnecessary.

\begin{definition}
\label{def:def_unnecessaryrepresentationcomplexity}
\textbf{(Unnecessary Representational Complexity)}
Representational complexity in \(R_0\) is unnecessary with respect to information scope Q when there exists an alternative representation \(R_1\) such that:
\begin{equation}
\text{Info}_Q(R_1) \equiv \text{Info}_Q(R_0)
\end{equation}
and
\begin{equation}
\text{ERC}(R_1) \prec \text{ERC}(R_0)
\end{equation}

\end{definition}
The existence of an information-equivalent representation with lower representational extent demonstrates that some of the representational extent of \(R_0\) was not necessary to represent the information in \(Q\).

Two recurring conditions can give rise to such reducibility. These conditions characterize properties of the existing representation and are not mutually exclusive.

\textbf{Redundancy} occurs when the same or overlapping enterprise information is represented through more artifacts than are required to preserve that information. The redundant artifacts may occur among Objects, Interactions, Behaviors, or Supporting Sources. Their removal or restructuring does not remove enterprise information because that information remains represented elsewhere in the simplified representation.

\textbf{Representational fragmentation} occurs when higher-order enterprise information is distributed across multiple lower-order structures that must be combined to reconstruct it. The individual structures may be valid and non-redundant; fragmentation concerns the distribution of the enterprise information across those structures.

These conditions may overlap. Fragmented information may also contain redundant representations, and reducing redundancy may also reduce fragmentation. Together, redundancy and representational fragmentation explain why representational reduction may be possible.

\subsection{Preservation Precedence}
Information preservation must be established independently of representational reduction. A decrease in ERC cannot compensate for failure to preserve an information requirement within \(Q\).
\begin{equation}
R_0 \equiv_Q R_1 \quad \text{before} \quad \text{ERC}(R_1) \prec \text{ERC}(R_0)
\end{equation}

Accordingly, a transformation that reduces ERC but fails any required preservation criterion constitutes information loss rather than ERS. Preservation therefore acts as an admissibility gate for subsequent classification of the transformation.

\subsection{Enterprise Representation Simplification Defined}
\begin{definition}
\label{def:def_ERS}
\textbf{(Enterprise Representation Simplification)}

A transformation \(T: R_0 \to R_1\) constitutes Enterprise Representation Simplification with respect to information scope \(Q\) when information preservation has first been established,
\begin{equation}
R_0 \equiv_Q R_1
\end{equation}

and the resulting representation satisfies the component-wise simplification condition:
\begin{equation}
\text{ERC}(R_1) \prec \text{ERC}(R_0)
\end{equation}
\end{definition}
Here, \(\prec\) denotes component-wise non-increase across ERC dimensions with at least one strict decrease. Preservation and ERC reduction are therefore separate conditions: preservation establishes admissibility, while ERC establishes whether the admissible transformation constitutes Enterprise Representation Simplification.

\subsection{Simplification Mechanisms}
ERS may be realized through consolidation, integration, abstraction, or combinations of these mechanisms. The mechanisms describe how the representation changes, whereas redundancy and representational fragmentation describe conditions that make information-preserving reduction possible.

\textbf{Consolidation} replaces multiple separately represented artifacts with fewer artifacts while preserving their scoped enterprise information. Consolidation may reduce redundancy, fragmentation, or both.

\textbf{Integration} combines complementary enterprise information distributed across multiple artifacts into fewer representations. It is particularly applicable to fragmentation but may also form part of a higher-order abstraction.

Abstraction restructures lower-order artifacts and their composition into fewer higher-order enterprise representations that directly express information previously obtained through the lower-order representation. Abstraction may simultaneously reduce representational fragmentation and redundancy.

For example, lower-order Objects and the Interactions required to compose them may be restructured as a higher-order Object:
\begin{equation}
\{O_1, O_2, O_3, I_{12}, I_{23}\} \to O^*
\end{equation}

provided that:
\begin{equation}
\text{Info}_Q(O^*) \equiv \text{Info}_Q(\{O_1, O_2, O_3, I_{12}, I_{23}\})
\end{equation}

Such a transformation may reduce both Object Complexity and Interaction Complexity because relationships previously required to reconstruct the enterprise concept no longer need to be exposed as separate representational artifacts.

The relationship between conditions and mechanisms is many-to-many. Fragmentation may be reduced through integration, consolidation, abstraction, or combinations thereof. Redundancy may be reduced through consolidation or abstraction. No particular mechanism defines ERS; the defining property is an information-equivalent reduction in representational extent.

\section[MeasuringChange]{Measuring Representation Change}
\subsection{ERC State Change}
Let \(T: R_0 \to R_1\) denote an architectural transformation between two representation states. The corresponding change in ERC is:
\begin{equation}
\Delta\text{ERC} = \text{ERC}(R_1) - \text{ERC}(R_0)
\end{equation}

\begin{equation}
\Delta\text{ERC} = (\Delta\text{OC}, \Delta\text{IC}, \Delta\text{BC}, \Delta\text{SC})
\end{equation}

where, for example, \(\Delta\text{OC} = \text{OC}(R_1) - \text{OC}(R_0)\), and equivalently for \(IC\), \(BC\), and \(SC\). The structured delta identifies where representational extent changed without imposing equivalence among dimensions.

\subsection{Strict ERC Reduction}
Define \(\text{ERC}(R_1) \prec \text{ERC}(R_0)\) when each component of \(R_1\) is less than or equal to the corresponding component of \(R_0\) and at least one inequality is strict:
\begin{equation}
\begin{aligned}
\text{OC}(R_1) &\le \text{OC}(R_0) \\
\text{IC}(R_1) &\le \text{IC}(R_0) \\
\text{BC}(R_1) &\le \text{BC}(R_0) \\
\text{SC}(R_1) &\le \text{SC}(R_0)
\end{aligned}
\end{equation}
Equivalently:
\begin{equation}
\Delta\text{ERC} \preccurlyeq 0, \quad \text{with} \quad \Delta\text{ERC} \neq 0
\end{equation}

For fixed \(Q\), information preservation combined with strict ERC reduction provides unambiguous evidence of representational simplification.

\subsection{Mixed Representation Transformation}
A transformation may reduce some ERC dimensions while increasing another, for example:
\begin{equation}
\Delta\text{OC} < 0, \quad \Delta\text{IC} < 0, \quad \Delta\text{BC} > 0
\end{equation}

In this case, the resulting ERC states are not component-wise ordered:
\begin{equation}
\text{ERC}(R_1) \nprec \text{ERC}(R_0)
\end{equation}

The transformation therefore does not satisfy the strict ERS definition, even when information is preserved. It may nevertheless be architecturally or economically desirable. Mixed representation transformations should be interpreted through their transformation semantics and independently observed consequences rather than through an aggregate ERC score.

This distinction preserves ERC's multidimensionality: the measure records that some forms of representational extent decreased while another increased, but does not impose a universal judgment that one type of reduction necessarily outweighs another type of increase.
\begin{equation}
\begin{split}
\text{Define } Q \longrightarrow 
\text{Define Preservation Criteria} \longrightarrow \\
\text{Verify Preservation} \longrightarrow 
\text{Measure } \Delta\text{ERC} \longrightarrow \\
\text{Classify Transformation}    
\end{split}
\end{equation}
Extension is different. When enterprise information scope expands, new Objects, Interactions, Behaviors, or Supporting Sources may become necessary. An increase in ERC can therefore represent legitimate extension rather than architectural deterioration. Simplification and extension may also occur simultaneously when new scope is introduced while existing representation is consolidated.

\subsection{Interpreting Representational Change}
ERC characterizes representational states and differences between them but does not independently identify the transformations responsible for those differences. Consolidation, decomposition, abstraction, generalization, enrichment, substitution, and extension can produce similar ERC changes. Conversely, meaningful transformations can occur without changing ERC. An Object may gain attributes, an Interaction may change meaning, a Behavior may be modified, or a Supporting Source may be replaced while the corresponding counts remain unchanged. Transformations may also have effects across dimensions; for example, consolidation of Objects may eliminate or replace Interactions and Behaviors that are no longer required. Changes in ERC must therefore be interpreted in relation to the transformations that produced them.

\section[designprinciples]{Design Principles for Enterprise Representation Simplification}
Enterprise Representation Simplification applies established principles of abstraction, generalization, separation of concerns, loose coupling, and information hiding to enterprise representation. The separation of Objects, Interactions, Behaviors, and Supporting Sources provides a basis for determining where representation should reside, where flexibility is required, and where simplification creates value. The objective is not minimum representation, but the appropriate placement of shared meaning and necessary distinction.  All simplification principles operate within the preservation boundary established by; abstraction, generalization, consolidation, or elimination are admissible only where the applicable information requirements remain preserved.
\subsection{Separating Representation by Concern}
Objects, Interactions, Behaviors, and Supporting Sources should remain distinguishable architectural concerns even when physically combined. This allows each form of representation to be generalized, simplified, governed, and changed independently where appropriate, rather than allowing implementation boundaries to determine the organization of enterprise meaning.

The separation also provides natural boundaries for assigning responsibility and measuring the costs and benefits associated with different forms of representation.

\subsection{Generalize the Foundation and Preserve Distinction at the Edge}
ERS favors abstraction and generalization where meaning is shared and specialization where requirements diverge. Common Objects and reusable Interactions can provide a comparatively stable foundation, while Behaviors express distinctions arising from rules, calculations, policies, constraints, and derived meaning. Further specialization can remain with individual consumers where it is not broadly reusable.

A distinction may therefore be necessary without being necessary throughout the shared representation. ERS seeks to place each distinction at the narrowest scope sufficient for its required use. Distinctions required broadly may belong within foundational representation; those required by particular behaviors, contexts, or consumers need not be propagated into representations used more broadly.

This placement of distinction allows flexibility to be introduced closer to where it is required while limiting unnecessary specialization and change within the shared representation. The principle is not that all variation belongs at the edge, but that the scope of a distinction should correspond to the scope of its information requirements.

\subsection{Measure Where Simplification Creates Value}
ERS need not strictly reduce every ERC dimension. Under the component-wise definition, some dimensions may remain unchanged while at least one decreases. Changes in Objects, Interactions, Behaviors, and Supporting Sources can be related independently to observed cost, utilization, dependency, change effort, task economics, and AI performance.

This enables organizations to determine where complexity resides, which representational concerns require greater stability or flexibility, and where simplification warrants investment. Over time, these observations can also inform responsibility, governance, and resource allocation.

These principles favor a generalized and stable shared foundation with differentiation introduced progressively toward Behavior and consumer use. ERS therefore seeks not simply to reduce representation, but to organize it so that necessary complexity and variability occur where they can be managed with the least unnecessary impact on the rest of the enterprise.

\subsection{Implications for Implementation Burden}
ERC measures representational and not implementation complexity.  ERS transformations therefore can affect the obligations required to implement and sustain a representation. Consolidation can reduce duplicated implementation, elimination can remove corresponding implementation obligations, generalization can enable reuse, and reduced fragmentation can decrease mappings, intermediaries, integrations, and dependencies.

These effects are not automatic. Abstraction may reduce exposed representational complexity while relocating complexity to supporting implementation mechanisms. ERS therefore distinguishes between complexity that is eliminated or consolidated and complexity that is merely displaced. When representational requirements are removed without being recreated elsewhere, ERS can reduce both representational and implementation burden, contributing to the economic effects considered below.

\section[costs]{Representational Obligations and Economic Consequences}
Enterprise representation is not economically passive. Objects, Interactions, Behaviors, and Supporting Sources create recurring obligations because they must be maintained, governed, secured, monitored, changed, enhanced, and consumed. ERS can therefore have economic consequences when it eliminates representational obligations while preserving the information required by the defined scope.

The economic argument does not require the assumption that every reduction in ERC reduces total cost, nor does it require assigning a universal monetary weight to an Object, Interaction, Behavior, or Source. Instead, the appropriate comparison is between the complete cost-bearing obligations of two information-equivalent representation states and the recurring task costs of consuming those states.

\subsection{Representation Management Obligations}
Let \(M(R)\) denote the set of recurring management obligations induced by representation R. These obligations may include maintenance, governance, security, provenance, metadata management, monitoring, testing, dependency management, documentation, change management, enhancement, and administration.

For the four ERC dimensions, recurring global representation cost may be decomposed as:
\begin{equation}
C_G(R) = C_O(R) + C_I(R) + C_B(R) + C_S(R)
\end{equation}

where \(C_O\), \(C_I\), \(C_B\), and \(C_S\) denote the aggregate recurring costs associated with the Objects, Interactions, Behaviors, and Supporting Sources of \(R\). These functions may include both component-specific and shared obligations and need not be linear in \(OC\), \(IC\), \(BC\), or \(SC\).

This distinction is important. A consolidated Object may individually require more effort than any one Object it replaces, but the relevant economic comparison is not between the consolidated Object and a single predecessor. It is between the complete obligation sets of \(R_0\) and \(R_1\). 

Obligations that continue in the simplified representation are not new costs merely because they are concentrated in a different component.
Representation management also creates dependencies at two distinct architectural boundaries. Internal representation dependencies arise among the artifacts of \(R\) itself. Consumer dependencies arise when systems, processes, or analytical frameworks rely on those artifacts. The two should be distinguished because ERS may reduce not only the dependency structure within a representation but also the dependency surface presented to its consumers.

Let \(D_R (R)\) denote the set of represented dependencies among ERC dimensions within \(R\). Let \(C\) denote the set of consumers of \(R\), and define the consumer dependency set as:
\begin{equation}
\begin{split}
D_R(R) = \{ (c, a) \mid &c \in C, \,\\
&a \in R,\\
&\text{ and } c \text{ depends on a} \}
\end{split}
\end{equation}

Consumers may include applications, APIs, reports, analytics, data products, ETL or integration processes, AI agents, retrieval-augmented generation (RAG) frameworks, semantic retrieval systems, and other mechanisms that consume the enterprise representation. A RAG framework, for example, must work with the enterprise representation available to it. When that representation is fragmented across multiple Sources, Objects, Interactions, Behaviors, and supporting structures, RAG must navigate that complexity to retrieve and assemble the information required for a task.

Define Consumer Dependency Count as a derived architectural measure:
\begin{equation}
\text{CDC}(R) = |D_C(R)|
\end{equation}

ERC measures the extent of the representation, not the number of consumers that depend on it. Two representations with identical ERC may therefore support very different numbers of consumer dependencies. Conversely, simplifying a representation may eliminate relatively few representational elements while reducing many downstream dependencies when those elements are shared across consumers.

\subsection{Obligation Elimination and Introduction}
Let \(E_t\) denote the set of recurring obligations eliminated by transformation \(T: R_0 \to R_1\), and let \(N_t\) denote the set of genuinely new recurring obligations introduced by the transformation. The recurring obligations cost effect attributable to the transformation can then be expressed as:
\begin{equation}
\Delta C_G(T) = \sum_{n \in N_t} c(n) - \sum_{e \in E_t} c(e)
\end{equation}

Unchanged obligations are present in both representation states and therefore cancel from the comparison. Consequently, recurring obligations cost decreases when the cost of eliminated obligations exceeds the cost of genuinely introduced obligations:
\begin{equation}
\sum_{e \in E_t} c(e) > \sum_{n \in N_t} c(n) \implies C_G(R_1) < C_G(R_0)
\end{equation}

The magnitude of the savings is determined empirically. ERS establishes only that eliminating a positive-cost recurring obligation, without recreating that obligation elsewhere, eliminates its associated recurring cost.

\subsection{Cost-Reduction Cases}
A finite set of recurring cost-reduction mechanisms follows directly from the representation categories and the operations required to manage them. These mechanisms identify cases in which ERS can eliminate specific cost-bearing obligations.

Source exposure reduction. Eliminating an exposed Supporting Source from a representation does not necessarily eliminate the underlying source, which may remain necessary for implementation. Source reduction therefore considers only the obligations and economic effects attributable to exposing the source within the measured representation rather than the full cost of operating it. These may include incremental licensing, user or identity administration, connectivity, access and security configuration, source-specific monitoring, metadata management, governance, and audit requirements. Exposure may also increase security and risk, including attack surface, even where these effects do not result in an immediately observable monetary cost.

For source-related recurring cost:
\begin{equation}
C_S(R) = G(R) + \sum_{s \in S(R)} C_s(s, R)
\end{equation}

where source-specific obligations \((C_s)\) represent costs attributable to individual Supporting Sources and \(G\) represents shared source-management obligations. If \(S(R_1) \subset S(R_0)\), the eliminated source-specific obligations have positive cost, and no equivalent obligations are introduced or transferred elsewhere within the representation, the corresponding recurring representation cost decreases.

This reduction applies only to the economic burden associated with the source's exposure within the representation. It does not imply elimination of the underlying source or of costs associated with its continued use in implementation.

\textbf{Redundant Object consolidation}. When multiple unnecessary or overlapping Objects are replaced by a smaller information-preserving set, Object-specific obligations that existed only because the redundant Objects were separately represented are eliminated. The relevant comparison is the aggregate cost of maintaining the information-equivalent Object representation before and after simplification, not the complexity of any single resulting Object.

\textbf{Interaction elimination}. When Object simplification eliminates intermediary joins, mappings, bridge structures, interfaces, or other represented Interactions, obligations associated exclusively with maintaining, testing, documenting, governing, or executing those Interactions are eliminated.

\textbf{Behavior consolidation}. When duplicated or distributed rules, calculations, policies, constraints, or operational semantics are consolidated without loss of required information, obligations associated exclusively with maintaining duplicated Behavior representations are eliminated. Making previously implicit behavior explicit may increase BC while still reducing obligations elsewhere; such mixed transformations must be evaluated by the complete before-and-after obligation sets.

\textbf{Dependency reduction.} ERS can reduce two classes of dependency. First, representation fragmentation can create internal dependencies required only to coordinate separately represented Objects, Interactions, Behaviors, or Sources. When simplification removes those representational structures, associated internal dependencies in \(D_R (R)\) may cease to exist. Second, consumers may depend directly on the representation artifacts that simplification consolidates, integrates, or abstracts. When multiple consumer-facing artifacts are replaced by fewer information-equivalent artifacts, corresponding relationships in \(D_C (R)\) may also be eliminated. Obligations associated exclusively with eliminated dependencies (e.g., configuration, mapping, access management, monitoring, testing, documentation, change-impact analysis, coordination, and remediation) are no longer necessary when they are neither transferred nor replaced by equivalent obligations.

The recurring cost associated with dependencies may be expressed as:
\begin{equation}
C_D(R) = \sum_{d \in D_R(R)} c_R(d) + \sum_{d \in D_C(R)} c_C(d)
\end{equation}

where \(c_R(d)\) and \(c_C(d)\) denote the recurring obligations attributable to an internal representation dependency and a consumer dependency, respectively. If ERS removes a dependency and no equivalent dependency is introduced, the positive recurring cost associated exclusively with that dependency is eliminated.

Consumer dependency reduction can amplify the economic effect of simplification through reuse. If an artifact a is depended upon by \(k\) consumers, then a representational restructuring that removes the need for those \(k\) direct dependencies can eliminate multiple consumer-specific obligations even though the underlying ERC reduction occurs only once. The magnitude of this effect remains empirical because dependency costs and consumer usage patterns vary, but the dependency relationships themselves are directly measurable.

\textbf{Change- and Enhancement-Surface Reduction}. Enterprise representation can affect the extent to which a business change or enhancement must propagate across the architecture. Even when a component ultimately requires no modification, determining its impact may require examination, validation, testing, or coordination. Simplification can reduce these obligations by reducing the number of representational artifacts across which a change must be considered.

For change or enhancement \(e\), let \(CS(e,R)\) denote the set of representation components that require examination, validation, or modification. Representation Change Surface (RCS) is then defined as:

\begin{equation}
\text{RCS}(e, R) = \operatorname{card}(\text{CS}(e, R))
\end{equation}

RCS is not a component of ERC. ERC characterizes the representation itself, whereas RCS characterizes the surface across which a particular change must be evaluated or propagated. Information-equivalent representations with similar ERC may therefore have different RCS values because dependencies and variability are organized differently.

ERS may reduce RCS through consolidation, abstraction, dependency reduction, or the localization of variability. In particular, where shared Objects and Interactions provide a stable foundation and necessary distinctions are localized within Behaviors or consumer-specific representation, changes to those distinctions may affect fewer representational components. For applicable classes of change, this effect can be evaluated empirically as:

\begin{equation}
\text{RCS}(e, R_1) < \text{RCS}(e, R_0)
\end{equation}

A reduction in RCS is not guaranteed by a reduction in ERC. RCS provides a separately observable measure of whether simplification has successfully reduced the propagation of change and its associated change-management obligations.

\subsection{Representation and Task Economics}
The recurring economic consequences of a representation can be considered at two levels. The first is Global representation cost which captures obligations associated with maintaining, governing, securing, operating, and changing the broader representation and does not include implementation costs independent of the representation. Let 
\begin{equation}
C_G(R)
\end{equation}

denote these recurring global obligations for representation \(R\).
Secondly, Task-level cost captures the recurring cost of consuming the representation to satisfy an information requirement. For task \(q\), let 
\begin{equation}
C_q (q,R)
\end{equation}

denote this cost. It may include retrieval, source access, context construction, relationship resolution, application of Behavior, inference, tool invocation, validation, coordination, and remediation. Where relevant, associated resource consumption, such as computational energy, may also be measured separately from monetary cost.

Task-level ERC identifies the minimum representational extent required by the task, but it does not determine its monetary cost. Simplification can reduce task cost when representational reduction eliminates required work rather than merely relocating it. Fewer Supporting Sources may reduce source access or cross-source coordination; fewer Interactions may eliminate intermediary relationship resolution; and fewer required Objects or Behaviors may reduce context construction, interpretation, or validation.
Where task \(q\) occurs\( F_q\) times during a period, its recurring workload contribution is:
\begin{equation}
C_q(q, R) = F_q \cdot C_q(q, R)
\end{equation}

Across a workload \(W\), recurring task cost can be aggregated across its constituent tasks. Consequently, modest per-task differences may become economically significant at enterprise scale.
\begin{equation}
C_Q(R) = \sum_{q \in Q} F_q \cdot C_q(q, R)
\end{equation}

For AI tasks, cost per attempt may not capture the complete economic effect when representation also affects the probability of a successful outcome. Let \(A(q,R)\) denote task accuracy and \(C_q^R (q,R)\) the expected representation-dependent cost of one attempt. Expected Cost per Successful Task (ECST) provides an optional accuracy-adjusted comparison:
\begin{equation}
\text{ECST}(q, R) = \frac{C_q^R(q, R)}{A(q, R)}
\end{equation}

Accuracy and ECST are empirical consequences rather than ERC measures. Representation-dependent attempt cost may include context construction, inference, validation, and expected remediation, while physical data execution can be treated separately when it is materially affected by database technology, data volume, caching, network topology, query optimization, or other implementation characteristics. Environmental resource consumption, such as the energy required to execute a task, may likewise be measured as a separate empirical consequence where relevant. Detailed decomposition and the inclusion of monetary, computational, or environmental measures are therefore analytical choices rather than required components of ERC.

\subsection{Transformation Value and Break-Even}
A representation with lower recurring obligations may still require substantial investment to create. Let 
\begin{equation}
C_T(R_0 \to R_1)
\end{equation}

denote the one-time cost of designing, implementing, validating, deploying, and transitioning from \(R_0 \to R_1\). Transformation cost is evaluated separately from recurring global and task-level costs.

Let \(H\) denote a time horizon measured in consistent operating periods. A compact comparison of retaining \(R_0\) and transforming to \(R_1\) is:
\begin{equation}
\text{TC}_H(R_0) = H \cdot [C_G(R_0) + C_Q(R_0)]
\end{equation}

\begin{equation}
\text{TC}_H(T) = C_T(R_0 \to R_1) + H \cdot [C_G(R_1) + C_Q(R_1)]
\end{equation}

Define the economic value of transformation \(T\) over horizon H as the difference between the cost of retaining the current representation and the cost of transforming and operating the alternative:
\begin{equation}
V_H(T) = \text{TC}_H(R_0) - \text{TC}_H(T)
\end{equation}

A positive value indicates that cumulative recurring savings have recovered the transformation investment over the selected horizon. Under constant periodic recurring savings \(S\), where:
\begin{equation}
S = [C_G(R_0) + C_Q(R_0)] - [C_G(R_1) + C_Q(R_1)]
\end{equation}

the break-even horizon, when \(S > 0\), is:
\begin{equation}
H^* = \frac{C_T(R_0 \to R_1)}{S}
\end{equation}

This separates three constructs that should not be conflated: \(\Delta\text{ERC}\) measures architectural change; changes in \(C_G\) and \(C_Q\) measure observed or estimated recurring economic consequences; and \(C_T\) measures the one-time investment required to achieve the transformation. A reduction in ERC therefore does not by itself establish positive economic value.

\subsection{Empirical Relationships and Prospective Decisions}
Repeated measurement can extend ERC from retrospective architectural comparison to empirical decision support. Because the ERC dimensions remain separate, organizations can observe whether changes in Objects, Interactions, Behaviors, or Supporting Sources are associated with different lifecycle, task, change, or performance consequences without imposing universal economic weights.

For an observed outcome Y, such as recurring cost, change effort, validation burden, or AI accuracy, the relationship may be represented conceptually as:
\begin{equation}
Y = f(\Delta\text{OC}, \Delta\text{IC}, \Delta\text{BC}, \Delta\text{SC}, X)
\end{equation}

where \(X\) represents workload, technology, organizational, and other environmental factors not captured by ERC.  Observed outcomes are empirical properties of the environment and cannot be inferred from ERC or its dimensions alone. 

Dimension-specific unit economics may also be estimated from observed transformations by relating attributable cost changes to changes in individual ERC dimensions. Such measures remain enterprise-, transformation-, and time-horizon-specific and do not assign a universal monetary value to an ERC unit.

Historical observations can subsequently inform prospective architectural alternatives. When alternative designs preserve the same required information scope, their expected changes in ERC can be compared with previously observed relationships among representational change, lifecycle obligations, task economics, change effort, and performance. ERC supplies architectural variables that may become useful predictors once their relationships with observed outcomes have been established.

Repeated application can therefore create a cumulative enterprise evidence base for resource allocation, governance priorities, transformation sequencing, and architecture forecasting. The economic role of ERC is consequently not to prescribe the least-cost architecture, but to make representational differences measurable so that their actual economic consequences can be learned and used in future decisions.

\subsection{What Can Increase Cost}
ERS does not imply that every economic component decreases. Cost can increase only where the transformation introduces a genuinely new obligation, shifts an existing obligation into a more costly form, or incurs a one-time transformation expense. Examples include a new abstraction or mediation service that must be operated, processing that is relocated rather than eliminated, or other persistent mechanisms required specifically by \(R_1\). These costs are captured in \(N_T\) rather than inferred merely from the fact that a resulting Object or Behavior is richer than one of its predecessors.

The transformation from \(R_0\) to \(R_1\) also incurs a one-time implementation cost. This cost is economically important but should be separated from the recurring cost of the resulting representation because it is not incurred indefinitely.

\section[AI]{Significance for Enterprise AI}
ERS affects AI reasoning by changing the representation from which task context is derived. Retrieval noise refers to representational conditions encountered during retrieval that complicate task interpretation or reasoning, including redundancy, fragmentation, ambiguity, competing representations, and unnecessary information.

Retrieval noise and task-level ERC are related but distinct. Task-level ERC measures the minimal representation required for a task, while retrieval noise reflects additional representational burden encountered in retrieving and interpreting it. ERS may reduce both by simplifying the underlying enterprise representation from which task context is derived.

\subsection{Enterprise Representation Simplification and Task Representation}
For task \(q\) under representation \(R\) let \(R_q \subseteq R\) denote the minimal sufficient representation required to satisfy \(q\). Task-level ERC characterizes the extent of this representation:
\begin{equation}
\begin{split}
\text{ERC}(q, R) = (&\text{OC}(q, R),\\
&\text{IC}(q, R),\\
&\text{BC}(q, R),\\
&\text{SC}(q, R))
\end{split}
\end{equation}

Retrieval may select representation beyond the minimum required for a task. For a representation state \(R_i\) and task \(q\), let \(R_q(q,R_i)\) denote the minimum representation required to satisfy \(q\), \(R_{\text{add}}(q,R_i)\) the additional representation selected for the task, and \(R_{\text{rem}}(q,R_i)\) the representation that remains unselected. Assuming retrieval includes the minimum required representation, these subsets collectively constitute the available representation:

\begin{equation}
 R_i = R_q(q,R_i)\cup R_{\text{add}}(q,R_i)\cup R_{\text{rem}}(q,R_i)  
\end{equation}

and the representation retrieved for the task is:

\begin{equation}
R_{\text{ret}}(q,R_i)=R_q(q,R_i)\cup R_{\text{add}}(q,R_i)
\end{equation}

Representation selected beyond the minimum task requirement is not necessarily noise. It may provide useful information, but it may also introduce redundancy, fragmentation, ambiguity, competing representations, unnecessary information, or other conditions that complicate interpretation or reasoning. Retrieval noise refers to these detrimental characteristics of the representation selected for the task.

ERS and retrieval therefore operate at different but related scopes. ERS transforms the enterprise representation available to consumers within information scope \(Q\), while retrieval selects representation for an individual task from the resulting representation:

\begin{equation}
R_0 \rightarrow R_1 \rightarrow R_{\text{ret}}(q,R_1)
\end{equation}

ERS may affect AI consumption in two representational ways. First, simplifying redundancy, fragmentation, ambiguity, and other unnecessary complexity in the underlying enterprise representation may limit the conditions that contribute to retrieval noise. This effect is not guaranteed because the representation ultimately encountered by the AI system also depends on the retrieval process. Second, transformation of the underlying representation may change the minimum representation required to satisfy the task, \(R_q(q,R_i)\), by changing how the required information, relationships, and Behaviors are organized and made available.

Second, ERS may simplify the minimum representation required to satisfy the task. The information required by \(q\) remains unchanged, but an information-equivalent representation may require fewer Objects, Interactions, Behaviors, or Supporting Sources to represent that information. This effect is reflected in task-level ERC:
\begin{equation}
\text{ERC}(q, R_1) \prec \text{ERC}(q, R_0)
\end{equation}

ERS can therefore affect AI consumption through three distinct mechanisms: (1) the representation from which task context is selected, potentially reducing conditions that contribute to retrieval noise; (2) the minimum representation required to satisfy the task, measured through task-level ERC; and (3) the architectural dependencies required to expose and manage that representation for the AI consumer.

These effects do not imply that ERS alone determines retrieval or reasoning performance. Retrieval effectiveness, model capability, prompting, representation quality, and intrinsic task difficulty may also affect observed outcomes.

\subsection{AI as a Representation Consumer}
AI frameworks are consumers of enterprise representation and therefore participate in the consumer dependency relationships defined above. An AI consumer may depend on Objects, Interactions, Behaviors, Supporting Sources, and other mechanisms required to access and use the representation. 

ERS may therefore affect AI architecture independently of reasoning performance. Replacing multiple consumer-facing artifacts with fewer information-equivalent representations can reduce the connectors, mappings, retrieval configurations, security policies, metadata structures, and change dependencies required to consume enterprise information. When simplified representations are shared, these reductions may extend across both AI and non-AI consumers.

\subsection{AI Reasoning Proposition}
Prior Text-to-SQL research provides empirical support for the proposition that schema complexity and information-equivalent changes in representation can affect AI reasoning performance (\cite{guo_towards_2019,kim_semantic-layer-mediated_2026,nascimento_text--sql_2026,shute_semantic_2026,wang_autolink_2026,zhang_case_2026,zhu_revisql_2026}), providing empirical motivation for testing whether the representational effects established above translate into improved AI performance.

ERS can therefore affect AI consumption through three distinct mechanisms: the unnecessary representation from which task context must be distinguished, the minimum representation that must be interpreted to satisfy the task, and the architectural dependencies required to expose and manage that representation for the AI consumer.

\begin{proposition}{AI Reasoning Proposition}
\label{prop:AI_reasoning}
For a fixed reasoning task \(q\), reductions in task-relative representational noise, task-level ERC, or both, produced through information-preserving ERS are hypothesized to improve AI reasoning performance when other experimental conditions are held constant.    

Conceptually:
\begin{equation}
\text{Performance}(q, R) = f(N(q, R), \text{ERC}(q, R), X_q)
\end{equation}

where \(X_q \)represents factors outside the measured representational effects, including model capability, prompting, retrieval effectiveness, representation quality, and intrinsic task difficulty.
\end{proposition}

\ref{prop:AI_reasoning} is empirical. ERS establishes measurable changes in the representation available to and required by an AI consumer; whether those changes improve reasoning performance must be established experimentally.

\section[example]{Illustrative Two-Phase Architectural Decision Example}
This example illustrates how ERS (ES), Enterprise Representation Complexity (ERC), and independently observed economic and performance measures can support both current-state comparison and prospective architectural decisions. Phase 1 compares a current representation with an information-equivalent simplified alternative. Phase 2 introduces a hypothetical real-time requirement and a change in the underlying source architecture to illustrate how the starting representation may affect the consequences of future change.

\textbf{Illustrative status}. All representation counts, costs, task volumes, accuracy values, dependency counts, and change-surface values are constructed scenario assumptions used only to demonstrate application of the proposed measures and relationships. They are not empirical findings or universal benchmarks. The distinction between recurring platform obligations and usage-dependent costs reflects common technology cost-modeling practices in which fixed or committed costs are considered separately from costs that vary with consumption  (\cite{noauthor_introduction_2025}). Accuracy is likewise treated as an independently measured performance outcome rather than inferred from architectural structure (\cite{keller_expanding_2026,nist_practices_2025}).

\subsection{Phase 1: Simplifying the Current Representation}
Assume historical Order information is available through a single enterprise lakehouse. Over time, analytics initiatives have created overlapping Order representations, Customer-Order views, project-specific joins, calculations, and consumer-specific artifacts. Although the information is exposed through a single Supporting Source, the representation contains substantial redundancy and representational fragmentation.

For illustration, the current representation \((R_0)\) contains 18 Objects, 14 Interactions, six Behaviors, and one Supporting Source:
\begin{equation}
\text{ERC}(R_0) = (18, 14, 6, 1)
\end{equation}

ERS reorganizes the same scoped information into three enterprise Objects, two Interactions, and one common Behavior. The lakehouse remains the single Supporting Source:
\begin{equation}
\text{ERC}(R_S) = (3, 2, 1, 1)
\end{equation}

Thus:
\begin{equation}
\begin{split}
\Delta\text{ERC} &= \text{ERC}(R_S) - \text{ERC}(R_0) \\
&= (-15, -12, -5, 0)
\end{split}
\end{equation}

Because the scoped information is preserved and no ERC dimension increases, the transformation \(T: R_0 \to R_S\) satisfies the strict ERS definition.

\subsubsection{Illustrative Phase 1 Consequences}
\begin{table}[htbp]
\centering
\caption{System Metrics and Financial Comparison between $R_0$ and $R_S$}
\label{tab:metrics_comparison}

\small
\setlength{\tabcolsep}{4pt}

\begin{tabular}{lll}
\toprule
\textbf{Measure} & \textbf{Current $R_0$} & \textbf{Simplified $R_S$} \\
\midrule
ERC & $(18, 14, 6, 1)$ & $(3, 2, 1, 1)$ \\
Consumer dependencies & 75 & 28 \\
One-time transformation & --- & \$150K \\
Global recurring obligations & \$70K/year & \$25K/year \\
Representation-dependent cost / attempt & \$0.10 & \$0.08 \\
Illustrative task accuracy & 75.0\% & 79.2\% \\
Annual task attempts & 3.0M & 3.0M \\
\bottomrule
\end{tabular}

\end{table}

The values illustrate different consequences that may be associated with the two representation states. The reduction from 75 to 28 consumer dependencies assumes that consolidation allows multiple consumers to reuse common Objects, Interactions, and Behaviors rather than maintaining dependencies on project-specific artifacts. The \$150,000 transformation cost represents the assumed one-time effort to design, implement, validate, and migrate to the simplified representation.

The recurring obligations represent costs that persist independently of individual task execution, such as maintenance, governance, metadata, mappings, and other representation-related operational responsibilities. The reduction from \$70,000 to \$25,000 illustrates the assumed effect of maintaining fewer consumer-facing artifacts and dependencies. These values are scenario assumptions rather than costs derived from ERC.

The representation-dependent costs of \$0.10 and \$0.08 per attempt represent assumed costs that vary with task execution, such as retrieval, processing, orchestration, or other consumption-sensitive activity. Separating recurring obligations from consumption-dependent unit costs is consistent with technology unit-economic approaches that distinguish baseline or committed costs from costs associated with workload usage (\cite{noauthor_introduction_2025}).
At three million annual task attempts, representation-dependent task cost is:
\begin{equation}
C_W(R_0) = 3{,}000{,}000(0.10) = \$300{,}000
\end{equation}
\begin{equation}
C_W(R_S) = 3{,}000{,}000(0.08) = \$240{,}000
\end{equation}
Total recurring cost under the illustrative assumptions is therefore:
\begin{equation}
C_R(R_0) = 70{,}000 + 300{,}000 = \$370{,}000
\end{equation}
\begin{equation}
C_R(R_S) = 25{,}000 + 240{,}000 = \$265{,}000
\end{equation}
\begin{equation}
S = 370{,}000 - 265{,}000 = \$105{,}000 \text{ per year}
\end{equation}
Against a \$150,000 one-time transformation cost, the illustrative break-even horizon is:
\begin{equation}
H^* = \frac{150{,}000}{105{,}000} \approx 1.43 \text{ years}
\end{equation}

The calculation does not imply that the ERC reduction intrinsically produces a \$105,000 annual saving. The scenario associates independently estimated economic obligations with two measurable representation states. In practice, these values would be estimated from enterprise-specific implementation, operating, and workload data.

\subsubsection{Accuracy-Adjusted Task Economics}
The illustrative accuracy values of 75.0\% and 79.2\% represent independently observed or experimentally estimated task performance under the two representations. They are intentionally not derived from ERC. AI evaluation guidance similarly treats accuracy and other performance characteristics as empirical measurements whose interpretation depends on the task and evaluation (\cite{keller_expanding_2026}).

Where representation affects both the cost of an attempt and the probability of success, Expected Cost per Successful Task (ECST) provides an accuracy-adjusted comparison:
\begin{equation}
\text{ECST}(q, R) = \frac{C_q^R(q, R)}{A(q, R)}
\end{equation}
For the Phase 1 current state:
\begin{equation}
\text{ECST}(q, R_0) = \frac{0.10}{0.75} \approx \$0.133
\end{equation}
For the simplified state:
\begin{equation}
\text{ECST}(q, R_S) = \frac{0.08}{0.792} \approx \$0.101
\end{equation}

If the enterprise requires three million successful outcomes, the expected representation-dependent cost is approximately \$400,000 for \(R_0\) and \$303,030 for \(R_S\). The distinction is important: ECST is a cost per successful outcome and should therefore be applied to required successful outcomes, not simply multiplied by an observed number of attempts. Accuracy remains an illustrative empirical consequence rather than a component of ERC.

\subsection{Phase 2: Prospective Real-Time Requirement}
Assume the enterprise subsequently requires real-time Order information from multiple operational systems. Meeting this requirement requires a change in the source architecture. Rather than relying on the historical lakehouse as the single source of consumer-facing Order information, the enterprise introduces a federation platform capable of accessing and integrating the operational sources in real time.

Phase 2 is prospective. The question is not simply whether federation should be adopted, but how introducing the same source capability may propagate from two different representational starting points.

Under the direct path, the real-time sources and their differences are exposed through extensions to the existing representation. The scenario assumes that this introduces six additional Objects, six Interactions, three Behaviors, and three additional exposed Supporting Sources:
\begin{equation}
\text{ERC}(R_E) = (24, 20, 9, 4)
\end{equation}

Under the simplified path, the federation platform replaces the lakehouse as the Supporting Source presented through the established enterprise representation. Source-specific schemas and integration differences are resolved behind that representation. The three Objects, two Interactions, and one Behavior therefore remain stable, and federation remains exposed as a single Supporting Source:
\begin{equation}
\text{ERC}(R_{SE}) = (3, 2, 1, 1)
\end{equation}

The simplified path therefore does not become progressively “more simplified” in Phase 2. Instead, the source implementation changes while the consumer-facing representation remains stable.

\subsubsection{Prospective Phase 2 Consequences}
\begin{table}[htbp]
\centering
\caption{Phase 2 values}
\label{tab:enhanced_metrics}

\small
\begin{tabular}{lll}
\toprule
\textbf{Measure} & \textbf{Enhanced $R_E$} & \textbf{Enhanced $R_{SE}$} \\
\midrule
Projected ERC & $(24, 20, 9, 4)$ & $(3, 2, 1, 1)$ \\
Consumer dependencies & 110 & 28 \\
Representation Change Surface & 12 & 0 \\
Phase 2 transformation cost & \$120K & \$80K \\
Global recurring obligations & \$100K/year & \$275K/year \\
Representation-dependent cost / attempt & \$0.18 & \$0.08 \\
Illustrative task accuracy & 65.0\% & 79.2\% \\
\bottomrule
\end{tabular}
\end{table}

The projected increase from 75 to 110 consumer dependencies on the direct path reflects the assumed propagation of the new sources and representation structures into existing consumer relationships. The simplified path remains at 28 because consumers continue to depend on the established enterprise representation rather than directly on the newly introduced operational sources.
The Representation Change Surface of 12 for \((R_E)\) represents the assumed number of existing consumer-facing representation elements requiring modification as the real-time capability is introduced. The value of zero for \((R_{SE})\) indicates that the existing consumer-facing Objects, Interactions, and Behaviors do not require modification; this does not imply that no implementation work occurs behind them.

The \$80,000 Phase 2 transformation cost for \((R_{SE})\) therefore represents the assumed work required to introduce the federation platform, connect the operational systems, and map them to the existing representation. The \$120,000 direct-path estimate assumes both source integration and changes propagated through the existing consumer-facing representation.

The increase in global recurring obligations from \$25,000 in Phase 1 to \$275,000 for \((R_{SE})\) is deliberate. It represents the assumed recurring cost of the federation capability, including platform licensing and associated operation, in addition to the remaining representation obligations. Technology architectures can combine substantial fixed or committed platform costs with usage-sensitive costs, making it important to evaluate both rather than assuming that the representation with lower structural complexity must have the lowest cost in every category (Capability, 2026).
Thus, the simplified alternative incurs a substantially higher fixed recurring obligation to obtain real-time federation capability while preserving its lower consumer-facing ERC and assumed \$0.08 representation-dependent cost per attempt

At three million annual task attempts:
\begin{equation}
C_W(R_E) = 3{,}000{,}000(0.18) = \$540{,}000
\end{equation}
\begin{equation}
C_W(R_{SE}) = 3{,}000{,}000(0.08) = \$240{,}000
\end{equation}
\begin{equation}
C_R(R_E) = 100{,}000 + 540{,}000 = \$640{,}000
\end{equation}
\begin{equation}
C_R(R_{SE}) = 275{,}000 + 240{,}000 = \$515{,}000
\end{equation}
\begin{equation}
S = 640{,}000 - 515{,}000 = \$125{,}000 \text{ per year}
\end{equation}

Under these assumptions, the simplified alternative has a \$125,000 annual recurring advantage despite its higher global recurring obligation. The result arises from the assumed combination of fixed platform cost, workload volume, and representation-dependent unit cost; it is not implied by ERC alone.

\begin{figure}
    \centering
    \includegraphics[width=1\linewidth]{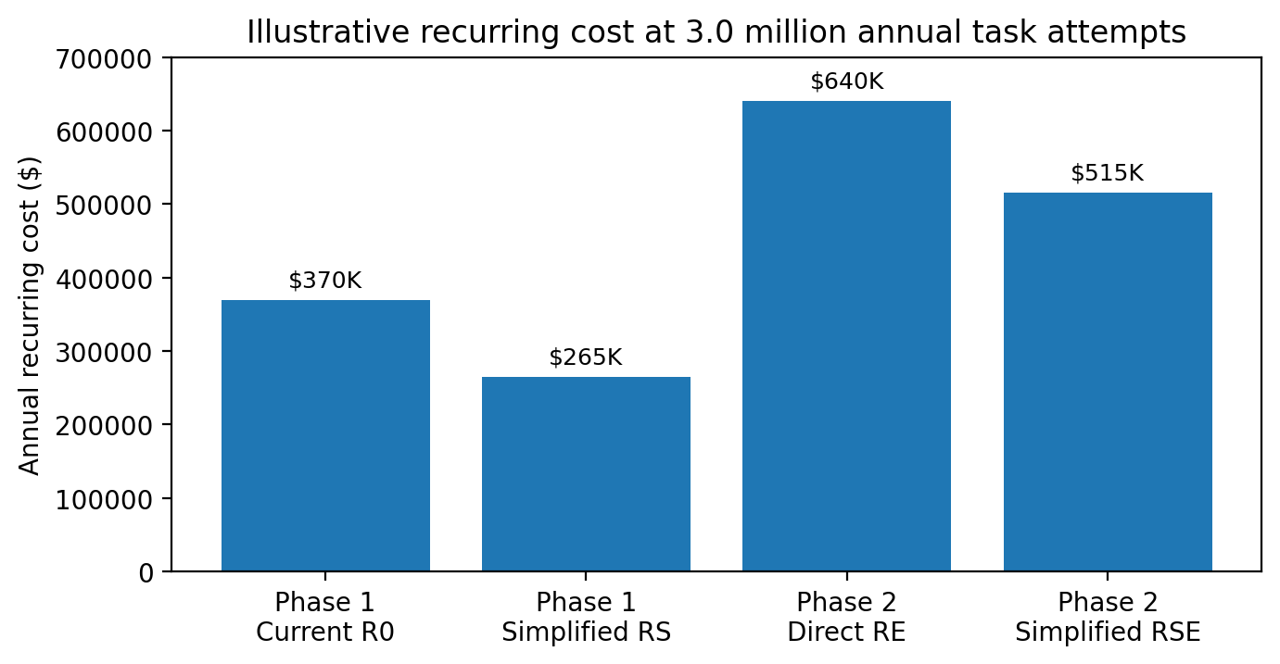}
    \caption{Illustrative recurring-cost comparison at 3.0 million annual task attempts}
    \label{fig:sample_1}
\end{figure}
\subsubsection{Phase 2 Accuracy-Adjusted Comparison}
\begin{equation}
\text{ECST}(q, R_E) = \frac{0.18}{0.65} \approx \$0.277
\end{equation}

\begin{equation}
\text{ECST}(q, R_{SE}) = \frac{0.08}{0.792} \approx \$0.101
\end{equation}

If three million successful outcomes are required, the expected representation-dependent cost is approximately \$830,769 for \(R_E\) and \$303,030 for \(R_{SE}\). 

The simplified ECST remains the same as in Phase 1 because the simplified representation, task cost, and illustrative accuracy are intentionally held constant across phases. The underlying source architecture, however, has changed from a lakehouse to a real-time federation platform.
\begin{figure}
    \centering
    \includegraphics[width=1\linewidth]{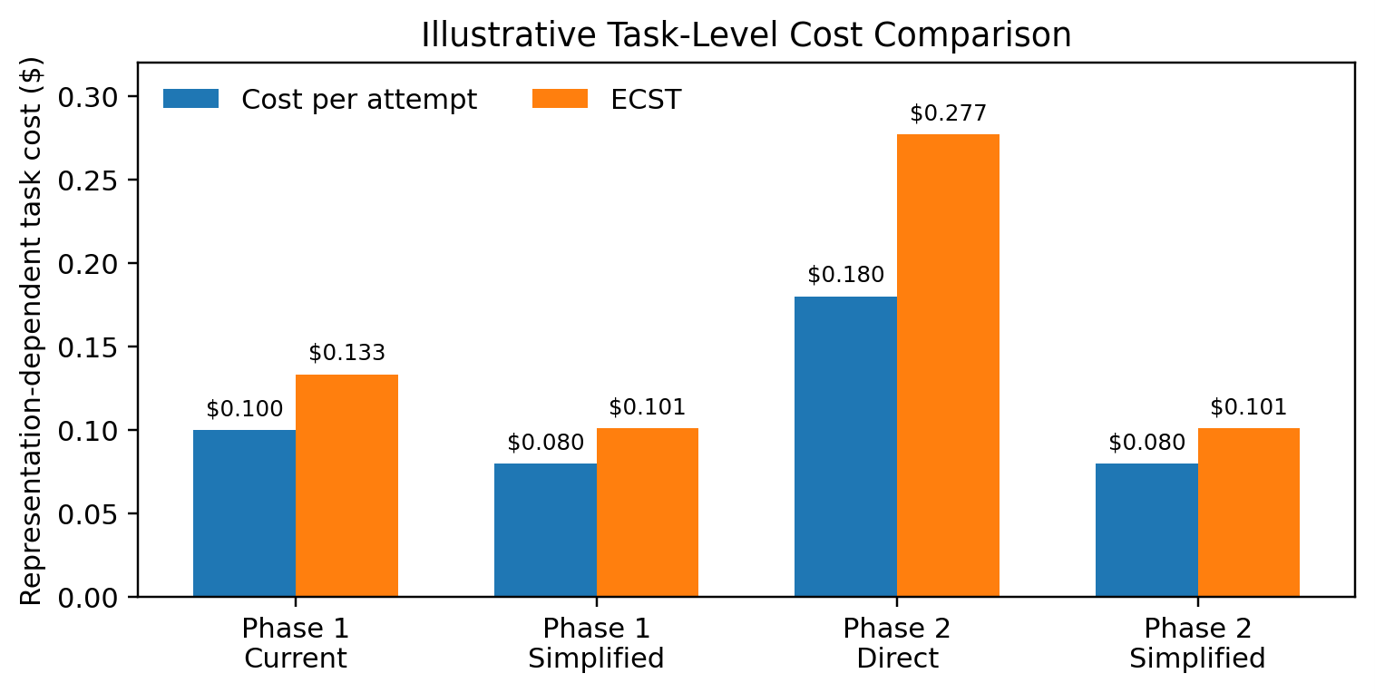}
    \caption{Illustrative task cost per attempt and Expected Cost per Successful Task (ECST).}
    \label{fig:sample_2}
\end{figure}
\subsubsection{Prospective Decision Support}
The central result of Phase 2 is not that one architecture universally wins. It is that the same future requirement can produce materially different projected consequences depending upon the representation from which the enterprise begins. For proposed enhancement \(e\) applied to starting representation \(R\), the enterprise can estimate:
\begin{equation}
\Delta\widehat{\text{ERC}}(e, R)
\end{equation}

\begin{equation}
\widehat{C}_T(e, R), \quad \widehat{C}_G(e, R), \quad \widehat{C}_q(e, R)
\end{equation}

using historical observations, scenario assumptions, or other forecasting methods. As observations accumulate across transformations, the empirical relationship between representational change and observed outcomes can increasingly be estimated from enterprise evidence:
\begin{equation}
Y = f(\Delta\text{OC}, \Delta\text{IC}, \Delta\text{BC}, \Delta\text{SC}, X)
\end{equation}

where \(X\) represents workload, technology, organizational, and other environmental factors not captured by ERC. ERC supplies the architectural variables; the economic and performance relationships remain empirical.

\subsection{Decision-Making Implications}
The example illustrates three uses of ERS and ERC. First, ERC makes the representational difference between current and simplified states explicit. Second, economic, performance, and other measurable outcomes can be associated with those states without being treated as intrinsic ERC values. Third, the resulting evidence can be carried forward to compare how future requirements may propagate from different architectural starting points.

The illustrative analysis considers monetary consequences only. Other measurable consequences of representation may be evaluated separately where relevant. For example, computational energy consumed during task execution can be compared across task-equivalent representations to determine whether differences in representational extent correspond to differences in environmental resource consumption. Such effects are empirical consequences of representation and are not implied by ERC reduction itself.

ERS therefore supports more than retrospective reduction. Repeated measurement can create an enterprise-specific evidence base for forecasting transformation effort, recurring obligations, task economics, change propagation, and performance under alternative representation designs.

\section[researchagenda]{Research Agenda}
The work establishes a research agenda for evaluating both the measurement properties of ERC and the outcomes associated with ERS. Four complementary areas are particularly relevant.

\subsection{Representation and Measurement Validation}
Controlled representation transformations can evaluate whether ERS produces measurable reductions in representation- and task-level ERC while preserving required information. Such studies can compare information-equivalent representation states and examine how alternative transformations affect ERC dimensions.

Measurement studies can additionally evaluate the reproducibility with which independent practitioners identify these representational categories. Inter-rater reliability and classification consistency would provide evidence concerning the operational reliability of ERC. Because the present work operationalizes the ERC complexity functions using cardinality, subsequent studies may also evaluate whether alternative category-specific functions provide greater explanatory power while preserving consistent comparison across representation states.

\subsection{AI Reasoning and Task-Level Effects}
AI reasoning studies can test Proposition \ref{prop:AI_reasoning} by comparing performance across information-equivalent representations with different task-level ERC values while holding model, prompting, questions, underlying information, and evaluation criteria constant. Text-to-SQL provides a particularly suitable experimental setting because representation structures, task requirements, generated queries, and execution results can be observed directly.

Task-level studies can further examine whether differences in \(ERC(q,R)\), or in its individual dimensions, are associated with retrieval effort, source access, context construction, inference cost, tool use, validation, remediation, and accuracy. This would allow empirical analysis to determine which forms of representational complexity are consequential for particular AI workloads rather than assuming equivalent effects across ERC dimensions.

\subsection{Lifecycle, Economic and Resource Effects}
Lifecycle studies can examine relationships between representation-level ERC and maintenance, governance, security, dependency management, change, and enhancement effort. Representation Unit Economics can extend this analysis by estimating dimension-specific unit and marginal costs associated with individual ERC dimensions. These studies can determine how changes in individual ERC dimensions relate to observed economic and/or resource effects. 

Task-level studies can similarly examine whether differences in representational extent correspond to differences in computational resource consumption. For AI-enabled workloads, this may include energy consumed during retrieval, context construction, inference, tool invocation, validation, and other task activities. Comparing resource consumption across task-equivalent representations could determine whether reductions in ERC are associated with lower energy per task or per successful task, and which representational dimensions contribute most strongly to observed differences. Such relationships require empirical measurement and are not implied by ERC reduction.

Changes in ERC can also be examined over time as enterprise representations evolve through simplification, extension, and other transformations. Repeated measurement can determine whether changes in representational complexity correspond to changes in lifecycle obligations, Representation Change Surface, operational effort, or other measurable outcomes. Representation Utilization can similarly be examined over time to determine how much of the maintained representation participates in observed enterprise workloads and whether changes in utilization provide useful evidence for architectural review.

\subsection{ERS and ERC for Decision Support}
Repeated measurement across tasks, workloads, and transformations provides an opportunity to determine whether relationships between ERC dimensions and observed outcomes persist over time. Such studies can examine whether changes in Objects, Interactions, Behaviors, or Supporting Sources exhibit sufficiently stable relationships with accuracy, representation-dependent task cost, lifecycle obligations, utilization, or enhancement effort to inform prospective estimates of architecture alternatives.

This research can also investigate whether accumulated ERC-based evidence changes architectural practice. Historical observations may reveal which representational dimensions contribute most strongly to particular outcomes, providing evidence for resource allocation, governance priorities, simplification opportunities, and representation-management practices. This would evaluate the use of ERS and ERC for individual representation transformations, and as the basis for an continuing organizational process of measurement, empirical learning, and architectural improvement.

\section[limitations]{Limitations}
ERC provides a basis for characterizing representational complexity but does not constitute a universal measure of enterprise complexity. The present work operationalizes its category-specific complexity functions using cardinality and therefore measures representational size rather than internal structural complexity, dependency topology, semantic or information quality, data volume, change frequency, or other characteristics that may distinguish structures having the same count. Alternative complexity functions may capture such characteristics but require separate empirical validation.

ERC also does not impose a universal scalar ordering across its dimensions. These dimensions represent different architectural properties, and weighting them would require assumptions about their relative significance that may vary across enterprises, technologies, tasks, and operating environments. Mixed transformations are therefore interpreted through information preservation and transformation semantics and not an aggregate complexity score.

ERC does not intrinsically determine utilization, Consumer Dependency Count (CDC), Representation Change Surface (RCS), monetary cost, or AI performance. Economic and resource relationships require enterprise-specific observations or explicit assumptions, while AI outcomes additionally depend on model capability, prompting, retrieval effectiveness, representation quality, task difficulty, and other factors outside ERC. Relationships between ERC and these outcomes are therefore empirical rather than intrinsic to the measure.

Representation Utilization is similarly workload-dependent. Low utilization may identify representation warranting further examination but does not establish that it is unnecessary, since infrequently used structures may support critical, regulatory, seasonal, exceptional, or future requirements.
ERS does not prescribe a universal preservation test across representation types; preservation criteria must be defined according to the information requirements within Q, and some requirements may require domain validation rather than purely mechanical equivalence testing.

Finally, ERC depends upon consistent identification of Objects, Interactions, Behaviors, and Supporting Sources and explicit definition of the information scope against which preservation is evaluated. Classification reproducibility, alternative complexity functions, and the stability of empirical relationships between ERC and observed outcomes therefore remain important areas for validation.

\section[conclusion]{Conclusion}
ERS treats representational complexity as an architectural condition that can be deliberately reduced rather than merely navigated. It is defined as an information-preserving transformation that reduces unnecessary representation relative to a defined enterprise information scope.
ERC makes representational complexity measurable through separate functions for each dimension:
\begin{equation}
\begin{split}
\text{ERC}(R) = (&\varphi_O(O(R)),\\
&\varphi_I(I(R)),\\ 
&\varphi_B(B(R)),\\ 
&\varphi_S(S(R)))   
\end{split}
\end{equation}

This paper operationalizes these functions using cardinality, providing a simple, observable, and technology-independent measure of representational size. The multidimensional structure preserves differences among representational categories rather than collapsing them into an unsupported scalar measure. Representation-level ERC characterizes the broader enterprise representation, while task-level ERC characterizes the minimum representation required to satisfy a particular information requirement.

ERS also provides a basis for examining the consequences of representational design. Representation creates recurring obligations associated with maintenance, governance, security, dependency management, change, and operation. At the task level, representation may affect context construction, inference, validation, and remediation. ERC does not assign economic cost, computational/resource burden, or reasoning performance to these conditions; it provides common architectural variables through which their relationships can be observed and compared. Representation Utilization and Representation Change Surface extend this analysis by distinguishing what is actually used and how broadly subsequent change propagates without incorporating those consequences into ERC itself.

For AI, ERS addresses a problem upstream of retrieval. Rather than accepting existing representational complexity and focusing only on selecting relevant context, simplification changes the representation from which context is derived. Reductions in task-level ERC may therefore reduce the representation an AI system must identify, relate, and interpret. Existing Text-to-SQL research supports the broader proposition that representational design can affect reasoning performance, while the specific relationships among ERC dimensions, accuracy, task economics, validation, and remediation remain empirical questions.

ERS therefore establishes an architectural objective, while ERC provides a measurement structure for evaluating progress toward it. ERS does not presume that fewer representational structures necessarily produce lower cost, greater agility, higher utilization, or better AI reasoning. Instead, it enables alternative representations to be measured consistently and their consequences evaluated empirically.

More broadly, repeated ERC measurement can turn individual architectural decisions into cumulative organizational evidence. Observations across representations, tasks, workloads, and transformations can reveal which representational conditions are consistently associated with utilization, reasoning performance, lifecycle obligations, and change effort. Those relationships can subsequently inform transformation priorities, resource allocation, and prospective architectural decisions. In this sense, ERS provides not only an approach to reducing unnecessary representation, but a basis for an enduring process of measurement, learning, and continuous representational improvement.

\bibliographystyle{unsrtnat}
\bibliography{my_references}  






\end{document}